%% file: acl_latex.tex
\documentclass[11pt]{article}

\usepackage{array}
\usepackage{amsmath}
\usepackage{amssymb}
\usepackage{booktabs}
\usepackage{multirow}
\usepackage{graphicx}
\usepackage[table]{xcolor}

\usepackage{tabularx}
\usepackage{xurl}
\newcolumntype{Z}{>{\raggedright\arraybackslash}X}

\newcommand{\sfclrepo}[1]{%
  \href{https://huggingface.co/datasets/#1}{\nolinkurl{#1}}%
}

\definecolor{oursblue}{RGB}{224,242,250}

\PassOptionsToPackage{hyperfootnotes=false}{hyperref}
\usepackage[preprint]{acl}

\usepackage{times}
\usepackage{latexsym}

\usepackage[T1]{fontenc}

\usepackage[utf8]{inputenc}

\usepackage{microtype}

\usepackage{inconsolata}

\usepackage{graphicx}

\title{Safe-FCL: Continual Risk Awareness and Fine-Grained Safety Aggregation for Federated Continual Learning of LLMs}

\author{
  \textbf{Shuaijie Ma}\textsuperscript{1}%
  \thanks{Equal contribution\quad
  \textsuperscript{\textdagger}Corresponding author}\quad
  \textbf{Yongkang Hu}\textsuperscript{1}\footnotemark[1]\quad
  \textbf{Yu Cheng}\textsuperscript{1,2}\quad
  \textbf{Jiuan Zhou}\textsuperscript{1,2}\\
  \textbf{Yi Xie}\textsuperscript{1}\quad
  \textbf{Yushuo Zhang}\textsuperscript{1}\quad
  \textbf{Zhaoxia Yin}\textsuperscript{1}\footnotemark[2]\\
  \textsuperscript{1}East China Normal University\\
  \textsuperscript{2}Shanghai Innovation Institute
}

\begin{document}
\maketitle

\begin{abstract}
Federated fine-tuning enables large language models (LLMs) to learn collaboratively from distributed client data, but malicious client updates can compromise global model safety. As federated tasks evolve, changing malicious update patterns challenge the ability of existing defenses to maintain reliable risk detection over time. To address this challenge, we propose \textbf{Safe-FCL}, a framework integrating continual risk awareness with fine-grained safety aggregation for federated continual learning of LLMs. Specifically, the \textbf{Continual Risk Awareness (CRA)} mechanism learns emerging risk patterns across sequential tasks and constrains detector updates using a protective subspace derived from historical features, mitigating forgetting of previously learned malicious patterns while adapting to new risks. To translate risk awareness into aggregation-level safety control, the \textbf{Fine-Grained Safety Aggregation (FGSA)} mechanism combines client-level risk assessments from CRA with module-level deviations from historical update behavior to adaptively regulate aggregation contributions, suppressing potentially malicious updates while minimizing unnecessary restrictions on benign updates. Extensive experiments demonstrate that Safe-FCL outperforms current SOTA methods, improving global-model Rule and MD-Judge scores in federated continual learning by 0.41\% and 0.64\%, respectively, while increasing average risk recognition accuracy by 2.01\%.

\end{abstract}

\section{Introduction}
\label{sec:introduction}

\begin{figure}[t]
    \centering
    \includegraphics[width=\columnwidth]{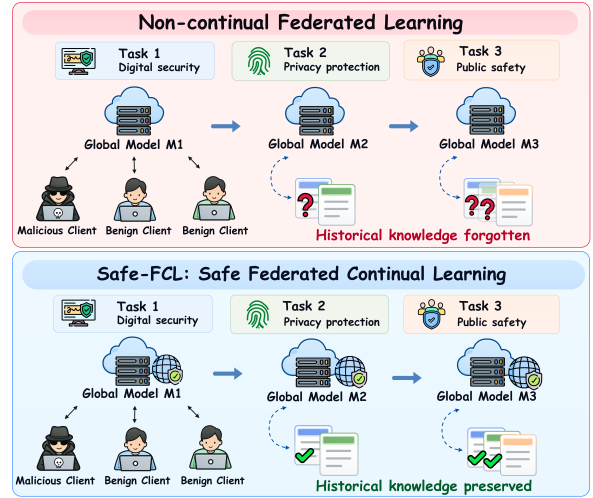}
    \caption{Comparison of non-continual federated learning and Safe-FCL.}
    \label{fig:fl_vs_fcl}
\end{figure}

Federated fine-tuning enables large language models (LLMs) to collaboratively learn from distributed client data without requiring the centralization of raw training data \citep{ye2024openfedllm}. However, malicious clients may compromise global model safety by manipulating their local model updates \citep{tao-etal-2026-safe}. As federated learning extends from single-task training to continual learning, models must acquire new knowledge while retaining previously learned capabilities \citep{pmlr-v139-yoon21b,pmlr-v306-qiu26w}. During this process, malicious update patterns may evolve as training tasks progress, requiring safety defenses not only to address current threats but also to maintain long-term effectiveness across sequential tasks.

Current studies have explored different strategies to defend against malicious client updates in federated learning. \citet{nguyen2022flame} proposed FLAME, which filters suspicious updates to mitigate their impact on the global model. \citet{Xu_2025_WACV} introduced LASA, employing layer-adaptive sparsified aggregation to improve robustness against malicious updates. More recently, \citet{tao-etal-2026-safe} developed Safe-FedLLM, which leverages LoRA update features to assess client risks and regulate aggregation contributions. \textbf{However, these methods do not explicitly address evolving malicious update patterns across sequential tasks.} Fixed risk detectors may miss emerging threats, while continual adaptation can lead to forgetting of previously learned risks. Moreover, client-level risk weighting lacks targeted control over suspicious updates across model modules.

To address these limitations, we propose \textbf{Safe-FCL}, which integrates continual risk awareness with fine-grained safety aggregation to enhance long-term safety in federated continual learning of LLMs. 
Specifically, \textbf{Continual Risk Awareness (CRA)} learns emerging malicious update patterns from model updates across sequential tasks. It constrains detector updates using historical protective subspaces to mitigate forgetting of previously learned risks. Building upon CRA, \textbf{Fine-Grained Safety Aggregation (FGSA)} combines client-level risk assessments with deviations from historical module update behavior. It then adaptively regulates module-level aggregation contributions through risk-aware gating. 
Together, these mechanisms leverage cross-task risk awareness to guide module-level aggregation decisions, thereby enhancing long-term safety in federated continual learning. Experimental results demonstrate that Safe-FCL outperforms current SOTA methods, improving global-model Rule and MD-Judge scores by 0.41\% and 0.64\%, respectively, while increasing continual risk recognition accuracy by 2.01\%.

The main contributions of this work are summarized as follows:

\begin{itemize}
    \item \textbf{Long-Term Safety Defense for Federated Continual Learning.}
    Safe-FCL is designed to mitigate long-term safety challenges posed by evolving malicious update patterns in federated continual learning of LLMs. By combining cross-task risk awareness with safety-aware aggregation, the framework aims to strengthen defenses across sequential tasks.

    \item \textbf{Continual Risk Awareness and Fine-Grained Safety Aggregation.}
    CRA constrains risk detector updates using protective subspaces derived from historical features to mitigate forgetting of previously acquired risk recognition capabilities. FGSA combines client safety scores with module-level historical deviations to adaptively regulate aggregation contributions through risk-aware gating.

    \item \textbf{Comprehensive Experimental Evaluation.}
    We evaluate Safe-FCL across sequential safety tasks and an unseen OOD task to examine its long-term safety and generalization capabilities. Compared with current SOTA methods, Safe-FCL improves global-model Rule and MD-Judge scores by 0.41\% and 0.64\%, respectively, while increasing risk detector average accuracy by 2.01\% in federated continual learning.

\end{itemize}

\section{Related Work}
\subsection{Federated Continual Learning}
Continual learning (CL) aims to learn sequential tasks while mitigating catastrophic forgetting. Representative strategies include parameter regularization \citep{kirkpatrick2017overcoming}, knowledge distillation \citep{li2016learning}, experience replay \citep{Rebuffi_2017_CVPR,buzzega2020dark}, and gradient constraints \citep{lopezpaz2017gradient,pmlr-v108-farajtabar20a,saha2021gpm}. Recent studies also explore continual adaptation for AI-generated image detection \citep{hu2026saido,zhang2026faced2cl}. Federated continual learning (FCL) further faces challenges from client data heterogeneity and update conflicts. Existing methods address these challenges through knowledge transfer and distillation-based compensation \citep{pmlr-v139-yoon21b,Dong_2022_CVPR}, feature anchoring and gradient correction \citep{Yu_2025_CVPR,Zhang_2025_ICCV}, and multi-expert collaboration \citep{yi2025multifcl}. As federated continual learning extends to LLMs, recent studies have explored federated continual instruction tuning \citep{Guo_2025_ICCV} and conflict-aware layer selection \citep{pmlr-v306-qiu26w}. However, these approaches primarily focus on knowledge retention and task adaptation, leaving long-term safety challenges posed by evolving malicious updates insufficiently explored.

\subsection{Safety Defenses for Federated LLMs}
While recent studies have investigated safety in evolving LLM-based systems \citep{cheng2026safecoevo,cheng2026tame}, federated learning presents an additional challenge: malicious client updates can compromise the global model. 
Existing defenses include Krum \citep{blanchard2017machine} for robust aggregation, FLTrust \citep{cao2021fltrust} for trust-based weighting, and FLAME \citep{nguyen2022flame} and FedREDefense \citep{pmlr-v235-xie24c} for anomaly filtering. LASA \citep{Xu_2025_WACV} further employs layer-adaptive filtering to enhance aggregation robustness. For federated LLM fine-tuning, Safe-FedLLM \citep{tao-etal-2026-safe} leverages LoRA update features to identify malicious behavior and mitigate harmful updates through risk-aware aggregation. However, as task distributions and malicious update patterns evolve, adapting to emerging threats while retaining historical risk recognition capabilities remains a key challenge for long-term safety.

\begin{figure*}[t]
    \centering
    \includegraphics[width=\textwidth]{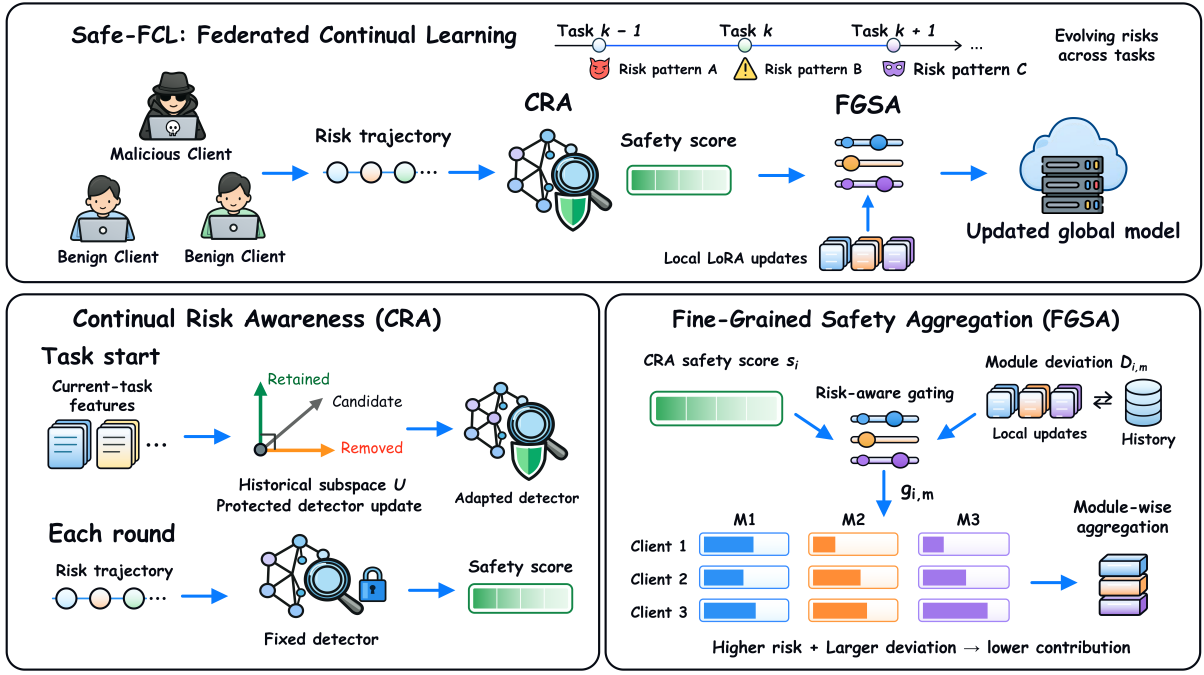}
    \caption{Overview of Safe-FCL. Continual Risk Awareness (CRA) adapts risk detection across sequential tasks while mitigating forgetting. Fine-Grained Safety Aggregation (FGSA) combines client-level risk assessments with module-level historical deviations to adaptively regulate aggregation contributions.}
    \label{fig:safe_fcl_framework}
\end{figure*}

\section{Methodology}
\subsection{Problem Setup and Framework Overview}
\label{sec:overview}

\noindent\textbf{Federated Continual Learning Setting.}
Consider a federated continual learning system with $N$ clients that sequentially learns $K$ tasks $\{\mathcal{T}_k\}_{k=1}^{K}$, with potentially heterogeneous data across clients and tasks. At communication round $r$ of task $\mathcal{T}_k$, the server selects a subset of clients $\mathcal{S}_{k,r}$, distributes the global LoRA adapter, and aggregates their locally fine-tuned updates. The global adapter parameters are inherited across tasks.

\noindent\textbf{Safety Threat Model and Objectives.}
In federated continual learning, malicious clients submit harmful model updates, and their behavior patterns may evolve across tasks. During aggregation, the server cannot access clients' raw data or know their true malicious identities, and must assess risks based on observable model updates. Therefore, this work aims to adapt to evolving malicious update patterns, preserve historical risk recognition capabilities, and suppress the impact of potentially malicious updates on the global model through fine-grained safety aggregation, thereby enhancing long-term safety in federated continual learning.

\noindent\textbf{Safe-FCL Framework Overview.}
Safe-FCL integrates continual risk awareness with fine-grained safety aggregation to support safety-aware federated fine-tuning across sequential tasks. During task-level risk detector adaptation, CRA learns emerging risk patterns from current-task update features and constrains detector updates using historical protective subspaces to mitigate forgetting. The detector then assesses client risks based on local update trajectories. FGSA further combines CRA-derived safety scores with historical deviations of actual module updates to adaptively regulate module-level aggregation contributions while maintaining module statistics online. When transitioning to a new task, Safe-FCL retains the global LoRA adapter parameters, risk detector, historical protection bases, and module-level statistics. This allows CRA to adapt to emerging risks while preserving historical risk knowledge, and enables FGSA to assess module updates using accumulated historical statistics.

\subsection{Continual Risk Awareness}
\label{sec:cra}

\begin{table}[t]
\centering
\caption{Key notations used in Safe-FCL.}
\label{tab:notation}
\footnotesize
\renewcommand{\arraystretch}{1.10}

\begin{tabular*}{\columnwidth}{
@{}p{0.27\columnwidth}
@{\extracolsep{\fill}}
p{0.70\columnwidth}@{}
}
\toprule
\textbf{Symbol} & \textbf{Description} \\
\midrule

$p_\phi(x)$ & Detector risk score \\
$U_{k,\ell}$ & Historical protection basis \\
$\widehat{\rho}_i$ & Suspicious-update proportion \\
$s_i$ & Client safety score \\

\addlinespace[2pt]
$\pi_i$ & Client aggregation weight \\
$E_{i,m}$ & Effective LoRA update \\
$z_{i,m},\,u_{i,m}$ & Magnitude and direction features \\
$\mu_m$ & Historical log-magnitude mean \\
$v_m$ & Historical log-magnitude variance \\
$d_m$ & Historical mean direction \\
$a_{i,m},\,b_{i,m}$ & Magnitude and directional deviations \\
$D_{i,m}$ & Overall module deviation \\
$g_{i,m}$ & Risk-aware retention coefficient \\

\bottomrule
\end{tabular*}
\end{table}

As federated continual learning progresses across sequential tasks, malicious client update patterns may evolve. Risk detectors trained on previous tasks may struggle to recognize emerging malicious update patterns, while continual adaptation to new risks may lead to forgetting of previously acquired risk recognition capabilities.

To address this challenge, we propose Continual Risk Awareness (CRA), which enables the risk detector to learn emerging malicious update patterns across sequential tasks while preserving historical risk knowledge. Specifically, CRA collects LoRA update trajectories from benign and malicious clients, extracts risk features from cumulative changes in selected LoRA parameters, and obtains risk scores using the detector $f_\phi$:

\begin{equation}
\label{eq:cra_risk_score}
p_\phi(x)=\sigma(f_\phi(x)).
\end{equation}

Here, $\phi$ denotes the detector parameters and $\sigma(\cdot)$ is the sigmoid function. Higher scores indicate greater predicted risk of malicious updates.

To preserve historical risk recognition capabilities, CRA identifies dominant feature directions from the detector's layerwise input activations associated with previous tasks. These directions are used to construct an orthonormal protection basis $U_{k,\ell}$ for each layer. To preserve previously acquired risk knowledge, CRA constrains detector updates along the protected directions, thereby mitigating interference with historical risk representations. During subsequent task adaptation, CRA uses this basis to constrain detector parameter updates, suppressing changes along important historical feature directions to reduce interference between new and previously learned risk knowledge. Let $\bar W_\ell$ denote the weights and biases of layer $\ell$, and $D_\ell=\bar W_\ell'-\bar W_\ell$ denote the parameter change produced by the optimizer. The constrained update is given by:

\begin{equation}
\label{eq:cra_projection}
\bar W_\ell^+
=\bar W_\ell+D_\ell
-(D_\ell U_{k,\ell})U_{k,\ell}^{\top}.
\end{equation}

This constraint suppresses parameter changes along important historical feature directions, thereby reducing interference with previously acquired risk recognition capabilities.

After detector training on the current task, its parameters $\phi$ remain fixed throughout the corresponding federated training rounds. CRA assesses client risks using local training trajectories separate from the actual updates used for aggregation. For client $i$ in the current communication round, CRA divides its local samples into $T$ equally sized effective batches, extracts update features $x_{i,u}$ at each step, and applies the risk scoring function $p_\phi(x)$ to identify suspicious updates. The suspicious-update proportion and client safety score are defined as:
\begin{equation}
\label{eq:cra_safety_score}
\begin{aligned}
\widehat{\rho}_i
&=\frac{1}{T}\sum_{u=1}^{T}
\mathbb{I}\left(p_\phi(x_{i,u})\geq\tau_{\mathrm{cls}}\right),\\
s_i&=(1-\widehat{\rho}_i)^\eta.
\end{aligned}
\end{equation}
Here, $\mathbb{I}(\cdot)$ is an indicator function that equals 1 when the risk score reaches the threshold $\tau_{\mathrm{cls}}$, indicating a suspicious update, and 0 otherwise. The parameter $\eta>0$ controls the risk suppression strength. A higher safety score $s_i$ indicates lower detected risk and serves as a client-level safety signal to guide subsequent FGSA aggregation decisions.


\subsection{Fine-Grained Safety Aggregation}
\label{sec:fgsa}

Existing safety aggregation methods based on client-level risk typically assign uniform weights to all model modules of the same client, limiting their ability to selectively control localized anomalous updates. To address this limitation, we propose Fine-Grained Safety Aggregation (FGSA), which adaptively regulates module-level aggregation contributions.

For task $\mathcal{T}_k$ at communication round $r$, let $\mathcal{S}$ denote the set of participating clients. FGSA adjusts client contributions based on the safety scores $s_i$ provided by CRA. If the average safety score of participating clients falls below a threshold $\tau_{\mathrm{skip}}$, indicating elevated overall detected risk, aggregation is skipped to prevent potentially harmful updates from affecting the global model, leaving the global parameters and historical module statistics unchanged. Otherwise, the base aggregation weights are computed as:
\begin{equation}
\pi_i=\frac{n_i s_i}{\sum_{j\in\mathcal{S}}n_j s_j},
\end{equation}
where $n_i$ denotes the number of local training samples at client $i$.

Building on client-level safety weighting, FGSA further examines module-specific updates to identify localized deviations from historical behavior. To this end, FGSA extracts magnitude and directional features from module updates. For LoRA module $m$, $E_{i,m}$ and $z_{i,m}$ are computed as:

\begin{equation}
\begin{aligned}
E_{i,m} &= \kappa(B_{i,m}A_{i,m}-B_mA_m),\\
z_{i,m} &= \log\max\left(
\frac{\|E_{i,m}\|_F}{\gamma_r},\epsilon
\right).
\end{aligned}
\end{equation}
Here, $A_m,B_m$ and $A_{i,m},B_{i,m}$ denote the global and local LoRA parameters, respectively. The terms $\kappa$, $\gamma_r$, and $\epsilon$ represent the LoRA scaling factor, current learning rate, and numerical stability constant. Meanwhile, FGSA extracts a normalized directional feature $u_{i,m}$ through a fixed low-dimensional projection to assess module deviations in both magnitude and direction.

Based on these features, FGSA compares current module updates with their historical behavior as follows:

\begin{equation}
\label{eq:module_deviation}
\begin{aligned}
a_{i,m} &= \max\left(0,\frac{z_{i,m}-\mu_m}{\sigma_m}-\delta_a\right),\\
b_{i,m} &= \max\left(0,1-\operatorname{cos}(d_m,u_{i,m})-\delta_b\right),\\
D_{i,m} &= \min(D_{\max},a_{i,m}+b_{i,m}).
\end{aligned}
\end{equation}

The historical magnitude statistics are maintained over logarithmic update magnitudes. The normalization scale is defined as $\sigma_m=\max(\sqrt{v_m},\sigma_{\min})$. The parameters $\delta_a$ and $\delta_b$ control the tolerated magnitude and directional deviations, while $\sigma_{\min}$ and $D_{\max}$ specify the lower normalization bound and upper deviation bound, respectively. The resulting $D_{i,m}$ provides a basis for subsequent risk-aware gating.

Since deviations from historical behavior do not necessarily indicate malicious activity, FGSA combines the CRA-derived $s_i$ with $D_{i,m}$ as follows:
\begin{equation}
g_{i,m}=\exp[-\lambda(1-s_i)D_{i,m}].
\end{equation}
Here, $\lambda>0$ controls the gating strength. When a client exhibits higher detected risk and its module update deviates substantially from historical behavior, $g_{i,m}$ decreases, strengthening the suppression of potentially harmful updates while avoiding excessive restrictions based solely on normal module variations.

Using these retention coefficients, FGSA performs differentiated safety aggregation across modules. For either LoRA factor $F\in\{A,B\}$ of module $m$, the aggregated parameter is:
\begin{equation}
F_m'=F_m+\sum_{i\in\mathcal{S}}
\pi_i g_{i,m}(F_{i,m}-F_m).
\end{equation}

To continuously maintain historical statistics, FGSA updates them after evaluating all participating clients, using only eligible updates with high safety scores, small historical deviations, and limited gating suppression. Let $\bar z_m$ and $\bar u_m$ denote the average logarithmic magnitude and mean direction of eligible observations, respectively. The historical statistics are updated as:
\begin{equation}
\label{eq:module_statistics}
\begin{aligned}
\mu_m' &= (1-\beta)\mu_m+\beta\bar z_m,\\
v_m' &= (1-\beta)\left[v_m+\beta(\bar z_m-\mu_m)^2\right],\\
d_m' &= (1-\beta)d_m+\beta\bar u_m.
\end{aligned}
\end{equation}
Here, $\mu_m'$, $v_m'$, and $d_m'$ denote the updated historical magnitude mean, variance estimate, and mean direction, respectively, while $\beta\in(0,1)$ is the statistical update rate. Historical module statistics are inherited across tasks, providing adaptive references for subsequent deviation assessment and allowing FGSA to suppress suspicious updates while reducing unnecessary interference with benign task learning.

\begin{table*}[t]
\centering
\caption{Rule and MD-Judge results (\%) across sequential tasks. Bold and underlined averages indicate the best and second-best results per phase and metric, respectively (ties for best are bold).}
\label{tab:updated_rule_mdjudge}

\setlength{\tabcolsep}{5pt}
\renewcommand{\arraystretch}{1.00}
\setlength{\aboverulesep}{1pt}
\setlength{\belowrulesep}{1pt}

\resizebox{\textwidth}{!}{%
\begin{tabular}{@{}l c|ccccc|ccccc@{}}
\toprule
\multirow{2}{*}{Method} & \multirow{2}{*}{Task}
& \multicolumn{5}{c|}{Rule}
& \multicolumn{5}{c}{MD-Judge} \\
\cmidrule(lr){3-7}\cmidrule(l){8-12}
& & Task 1 & Task 2 & Task 3 & Task 4 & Avg.
  & Task 1 & Task 2 & Task 3 & Task 4 & Avg. \\
\midrule

\multirow{4}{*}{FedAvg}
& T1 & 46.62 & -- & -- & -- & 46.62 & 41.26 & -- & -- & -- & 41.26 \\
& T2 & 46.85 & 72.49 & -- & -- & 59.67 & 41.72 & 77.16 & -- & -- & 59.44 \\
& T3 & 36.60 & 66.67 & 33.33 & -- & 45.53 & 33.57 & 65.73 & 29.37 & -- & 42.89 \\
& T4 & 43.12 & 72.03 & 42.19 & 57.81 & 53.79 & 40.33 & 70.16 & 38.93 & 61.07 & 52.62 \\
\midrule

\multirow{4}{*}{FLAME}
& T1 & 64.34 & -- & -- & -- & 64.34 & 59.21 & -- & -- & -- & 59.21 \\
& T2 & 60.37 & 79.25 & -- & -- & 69.81 & 56.88 & 81.12 & -- & -- & 69.00 \\
& T3 & 46.15 & 71.33 & 36.13 & -- & 51.20 & 41.49 & 72.73 & 32.40 & -- & 48.87 \\
& T4 & 46.62 & 72.96 & 44.99 & 72.73 & 59.33 & 42.19 & 73.89 & 40.09 & 76.92 & 58.27 \\
\midrule

\multirow{4}{*}{LASA}
& T1 & 55.94 & -- & -- & -- & 55.94 & 48.95 & -- & -- & -- & 48.95 \\
& T2 & 55.71 & 78.32 & -- & -- & 67.02 & 53.85 & 82.98 & -- & -- & 68.42 \\
& T3 & 39.63 & 74.36 & 34.03 & -- & 49.34 & 35.20 & 74.36 & 29.60 & -- & 46.39 \\
& T4 & 47.09 & 77.16 & 44.29 & 61.77 & 57.58 & 42.19 & 77.86 & 36.36 & 66.43 & 55.71 \\
\midrule

\multirow{4}{*}{Safe-FedLLM-Step}
& T1 & 84.38 & -- & -- & -- & \underline{84.38} & 85.31 & -- & -- & -- & \underline{85.31} \\
& T2 & 92.07 & 87.88 & -- & -- & 89.98 & 96.04 & 90.91 & -- & -- & 93.48 \\
& T3 & 47.55 & 76.92 & 44.06 & -- & 56.18 & 43.12 & 77.62 & 38.69 & -- & 53.14 \\
& T4 & 92.54 & 90.68 & 88.81 & 91.14 & 90.79 & 94.17 & 92.77 & 89.04 & 98.60 & 93.65 \\
\midrule

\multirow{4}{*}{Safe-FedLLM-Client}
& T1 & 62.47 & -- & -- & -- & 62.47 & 57.58 & -- & -- & -- & 57.58 \\
& T2 & 61.54 & 82.52 & -- & -- & 72.03 & 57.58 & 85.78 & -- & -- & 71.68 \\
& T3 & 41.96 & 73.43 & 35.43 & -- & 50.27 & 37.53 & 73.43 & 31.47 & -- & 47.48 \\
& T4 & 68.76 & 83.45 & 65.50 & 80.19 & 74.48 & 65.03 & 85.78 & 63.17 & 85.78 & 74.94 \\
\midrule

\multirow{4}{*}{Safe-FedLLM-Shadow}
& T1 & 53.38 & -- & -- & -- & 53.38 & 48.25 & -- & -- & -- & 48.25 \\
& T2 & 89.98 & 91.14 & -- & -- & \underline{90.56} & 96.04 & 94.17 & -- & -- & \underline{95.11} \\
& T3 & 79.72 & 90.91 & 70.86 & -- & \underline{80.50} & 84.38 & 90.44 & 75.29 & -- & \underline{83.37} \\
& T4 & 92.77 & 93.47 & 90.68 & 91.61 & \underline{92.13} & 93.94 & 96.04 & 92.54 & 98.83 & \underline{95.34} \\
\midrule

\rowcolor{oursblue}
& T1 & 90.44 & -- & -- & -- & \textbf{90.44} & 92.31 & -- & -- & -- & \textbf{92.31} \\

\rowcolor{oursblue}
& T2 & 94.41 & 92.07 & -- & -- & \textbf{93.24} & 97.67 & 95.57 & -- & -- & \textbf{96.62} \\

\rowcolor{oursblue}
& T3 & 85.55 & 91.61 & 79.25 & -- & \textbf{85.47} & 88.81 & 91.84 & 82.05 & -- & \textbf{87.57} \\

\rowcolor{oursblue}
\multirow{-4}{*}{\textbf{Safe-FCL (Ours)}}
& T4 & 93.71 & 92.31 & 92.31 & 91.84 & \textbf{92.54} & 95.10 & 95.10 & 94.64 & 99.07 & \textbf{95.98} \\

\bottomrule
\end{tabular}%
}
\end{table*}

\begin{table*}[t]
\centering
\caption{Continual risk recognition performance (\%) under Protocols 1 and 2. Bold and underlined values indicate the best and second-best results per column, respectively (ties for best are bolded).}

\label{tab:aa_af_ood_updated}

\setlength{\tabcolsep}{6pt}
\renewcommand{\arraystretch}{1.15}

\resizebox{\textwidth}{!}{%
\begin{tabular}{@{}lccccccccc@{}}
\toprule

\multirow{2}{*}{Method}
 & \multicolumn{1}{c}{Task1}
 & \multicolumn{2}{c}{Task2}
 & \multicolumn{2}{c}{Task3}
 & \multicolumn{2}{c}{Task4}
 & \multirow{2}{*}{New ACC $\uparrow$}
 & \multirow{2}{*}{\shortstack{OOD Task\\ACC $\uparrow$}} \\

\cmidrule(lr){2-2}
\cmidrule(lr){3-4}
\cmidrule(lr){5-6}
\cmidrule(lr){7-8}

 & AA $\uparrow$
 & AA $\uparrow$ & AF $\downarrow$
 & AA $\uparrow$ & AF $\downarrow$
 & AA $\uparrow$ & AF $\downarrow$
 & & \\

\midrule

Safe-FedLLM-Step
 & 91.17
 & 92.59 & -2.33
 & 81.28 & 6.17
 & 87.04 & \textbf{-0.50}
 & 86.09 & 90.33 \\

Safe-FedLLM-Client
 & 88.33
 & 86.67 & 1.66
 & 91.11 & \textbf{-9.17}
 & \underline{92.50} & 1.11
 & 88.75 & 91.67 \\

Safe-FedLLM-Shadow
 & 75.30
 & 89.35 & \textbf{-6.43}
 & 85.50 & 7.85
 & 92.25 & \underline{1.07}
 & 91.44 & 77.17 \\

\midrule

Safe-FCL w/o CRA
 & \underline{92.60}
 & \underline{98.00} & \underline{-4.10}
 & \underline{93.13} & 7.15
 & 92.39 & 8.04
 & \underline{97.40} & \underline{96.30} \\

\rowcolor{oursblue}
\textbf{Safe-FCL (Ours)}
 & \textbf{94.73}
 & \textbf{98.85} & -3.44
 & \textbf{96.70} & \underline{3.77}
 & \textbf{94.51} & 6.53
 & \textbf{98.55} & \textbf{96.60} \\

\bottomrule
\end{tabular}%
}
\end{table*}

\begin{table*}[t]
\centering

\caption{Ablation results for global-model safety (\%). Task 1--4 report average Rule and MD-Judge scores across all tasks learned by each stage. OOD reports global-model safety after continued federated fine-tuning on the unseen task with the risk detector fixed. Bold and underlined values indicate the best and second-best results per column, respectively (ties for best are bolded).}

\label{tab:safe_fcl_ablation_ood}

\setlength{\tabcolsep}{6pt}
\renewcommand{\arraystretch}{1.15}

\resizebox{\textwidth}{!}{%
\begin{tabular}{@{}l*{5}{cc}@{}}
\toprule

\multirow{2}{*}{Method}
 & \multicolumn{2}{c}{Task1}
 & \multicolumn{2}{c}{Task2}
 & \multicolumn{2}{c}{Task3}
 & \multicolumn{2}{c}{Task4}
 & \multicolumn{2}{c}{OOD} \\

\cmidrule(lr){2-3}
\cmidrule(lr){4-5}
\cmidrule(lr){6-7}
\cmidrule(lr){8-9}
\cmidrule(l){10-11}

 & Rule & MD-Judge
 & Rule & MD-Judge
 & Rule & MD-Judge
 & Rule & MD-Judge
 & Rule & MD-Judge \\

\midrule

Safe-FCL w/o FGSA
 & \textbf{90.68} & \textbf{92.54}
 & \underline{93.01} & \textbf{96.74}
 & 84.46 & \underline{86.95}
 & \textbf{92.78} & 95.63
 & 88.58 & 93.01 \\

Safe-FCL w/o CRA
 & 89.74 & 91.61
 & 91.84 & 95.92
 & \underline{84.62} & \textbf{87.57}
 & 91.84 & \underline{95.69}
 & \underline{90.91} & \underline{96.27} \\

\midrule

\rowcolor{oursblue}
\textbf{Safe-FCL (Ours)}
 & \underline{90.44} & \underline{92.31}
 & \textbf{93.24} & \underline{96.62}
 & \textbf{85.47} & \textbf{87.57}
 & \underline{92.54} & \textbf{95.98}
 & \textbf{95.57} & \textbf{96.97} \\

\bottomrule
\end{tabular}%
}
\end{table*}

\section{Experiments}
\label{sec:experiments}

\subsection{Experimental Setup}
\label{sec:experimental_setup}

\noindent\textbf{Datasets and Task Protocol.}
We construct five safety-related tasks covering different scenarios, following the sequence $\mathcal{T}=$ \{physical, public, and material safety; privacy and information protection; property, identity, and information deception; digital security; discrimination and interpersonal harm\}. Each task contains 5,000 training samples, 714 validation samples, and 1,429 test samples, with a benign-to-harmful sample ratio of 7:3. Further details on dataset sources, task construction, and data partitioning are provided in Appendices~\ref{app:dataset-sources} and~\ref{app:data}. To evaluate the continual learning capability and risk recognition generalization of Safe-FCL, we establish two experimental protocols.

\noindent\textbf{Protocol 1.}
Under the federated continual learning setting, the system sequentially learns the first four tasks. After each learning stage, global-model safety is evaluated on all tasks learned so far, together with the continual risk recognition performance of the detector.

\noindent\textbf{Protocol 2.}
To evaluate the Global-model and risk detector's generalization to an unseen domain, we keep the detector parameters learned during the federated continual learning stages fixed on the OOD task, while federated fine-tuning of the global model and FGSA historical-statistics updates continue.

\noindent\textbf{Implementation Details.} We adopt Llama-3.1-8B \citep{grattafiori2024llama} as the backbone model and employ LoRA \citep{hu2022lora} for parameter-efficient fine-tuning. The federation consists of 10 clients, including 3 malicious clients, with 3 clients randomly selected per communication round. Each task is trained for 100 communication rounds, and each client performs 10 local optimization steps per round.

\noindent\textbf{Baselines and Evaluation Metrics.}
We compare Safe-FCL with FedAvg \citep{pmlr-v54-mcmahan17a}, FLAME \citep{nguyen2022flame}, LASA \citep{Xu_2025_WACV}, and the Step-Level, Client-Level, and Shadow-Level variants of Safe-FedLLM \citep{tao-etal-2026-safe}. Global-model safety is evaluated using Rule \citep{tao-etal-2026-safe} and MD-Judge \citep{li-etal-2024-salad}, which assess rule-based violations and semantic safety of generated responses, respectively. Continual risk recognition is measured by Average Accuracy (AA), Average Forgetting (AF), and New-task Accuracy (New ACC). AA measures average detection accuracy across learned tasks, AF quantifies forgetting of historical risk recognition, and New ACC averages current-task accuracy immediately after each adaptation stage, reflecting adaptation to emerging risks. Federated training settings are described in the Implementation Details paragraph of Section~\ref{sec:experimental_setup}, and metric definitions are provided in Appendix~\ref{app:metrics}.


\begin{figure}[t]
  \centering
  \includegraphics[width=\columnwidth]{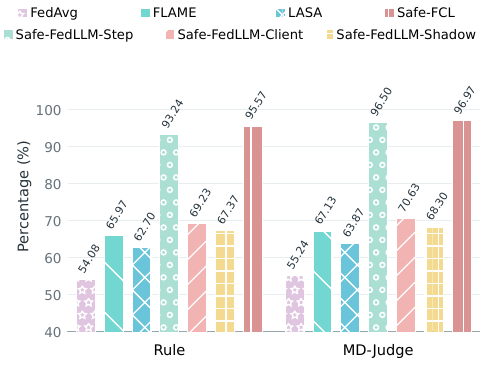}
  \caption{OOD generalization of federated global models. Rule and MD-Judge values are percentages.}
  \label{fig:ood-sequence2}
\end{figure}

\subsection{Safety Performance in Federated Continual Learning}
\label{sec:main_results}

Global-model safety is evaluated on the federated continual learning task sequence defined in Protocol 1. For fair comparison, all methods follow the same task order and federated training settings. As shown in Table~\ref{tab:updated_rule_mdjudge}, Safe-FCL achieves the highest average Rule and MD-Judge scores across all learning stages, demonstrating strong long-term safety defense capabilities. Specifically, upon completing federated continual learning, Safe-FCL achieves average Rule and MD-Judge scores of 92.54\% and 95.98\%, outperforming the current SOTA method by 0.41 and 0.64 percentage points, respectively. 

This performance reflects the complementary roles of risk recognition and aggregation control. As malicious update patterns evolve, CRA adapts client risk assessments while limiting interference with previously learned risk knowledge. FGSA further combines these assessments with historical magnitude and directional deviations to regulate individual LoRA modules. Rather than applying the same risk-based weight to every module of a client, this approach enables selective suppression of suspicious contributions, helping maintain global-model safety across changing tasks.

\subsection{Continual Risk Recognition Performance}
\label{sec:continual_risk_recognition}

The continual learning performance of the risk detector is evaluated on the task sequence defined in Protocol 1. For fair comparison, all methods follow the same task order and evaluation settings. As shown in Table~\ref{tab:aa_af_ood_updated}, Safe-FCL achieves the highest Average Accuracy across all learning stages, demonstrating strong continual risk recognition capabilities. Specifically, upon completing federated continual learning, Safe-FCL improves AA by 2.01 percentage points over the current SOTA method. In terms of adaptation to new tasks, Safe-FCL improves New ACC by 7.11 percentage points over the current SOTA method. Furthermore, under Protocol 2, Safe-FCL achieves an OOD Task ACC of 96.60\%, outperforming the strongest baseline by 4.93 percentage points.

These results demonstrate that Safe-FCL not only adapts to evolving malicious update patterns across sequential tasks but also preserves historical risk recognition capabilities, achieving a favorable balance between stability and plasticity. This benefit mainly stems from CRA, which constrains detector updates using historical orthonormal protection bases to reduce interference between new and previously learned risk knowledge. Moreover, Safe-FCL maintains high risk recognition accuracy on the OOD task without retraining the detector, indicating that CRA not only preserves the ability to recognize historical malicious update patterns but also enables learned risk discrimination capabilities to generalize to unseen task domains.

\subsection{OOD Generalization of Global-Model Safety}
\label{sec:ood_global_safety}
Adapting to an unseen safety domain introduces changes in client update behavior, posing additional challenges for safety-aware aggregation. As shown in Figure~\ref{fig:ood-sequence2}, Safe-FCL outperforms current SOTA methods by 2.33\% in Rule and 0.47\% in MD-Judge. The larger improvement in Rule suggests a stronger ability to suppress explicit safety violations, while the MD-Judge gain indicates that semantic-level safety is also preserved during adaptation.

On the unseen task, CRA leverages previously learned risk recognition capabilities to assess participating clients without additional detector training, demonstrating strong generalization performance. FGSA further combines these assessments with historical module-level deviations to selectively suppress suspicious updates rather than applying uniform restrictions across all modules. Meanwhile, by updating historical statistics using eligible observations, FGSA can adapt to shifts in benign update behavior. This adaptive control helps maintain global-model safety as federated fine-tuning continues on the unseen task.

\subsection{Ablation Study}
\label{sec:ablation_study}

Component ablations under the same experimental setting as Protocol 1 allow us to assess the effectiveness of CRA and FGSA. As shown in Table~\ref{tab:safe_fcl_ablation_ood}, the full Safe-FCL improves the final Rule and MD-Judge scores by 0.70 and 0.29 percentage points, respectively, compared with the variant without CRA. This suggests that continually adapting risk recognition helps maintain reliable safety aggregation as task distributions evolve. FGSA provides complementary benefits by regulating safety-critical module updates, with particularly noticeable improvements after the third task. However, its effects are not uniformly positive across all metrics at the final stage, indicating that module-level safety control involves trade-offs between different safety evaluation criteria rather than guaranteeing consistent gains.

The detector comparisons in Table~\ref{tab:aa_af_ood_updated} further clarify the role of CRA in preserving historical risk knowledge. Compared with the variant without CRA, Safe-FCL improves final AA by 2.12 percentage points while reducing AF by 1.51 percentage points. Improvements in New ACC and OOD Task ACC further suggest that protecting historical risk representations does not necessarily compromise adaptation to emerging malicious update patterns. Instead, CRA helps balance knowledge retention and plasticity, allowing the detector to recognize newly encountered risks while limiting interference with previously acquired capabilities. Taken together, these results highlight the complementary roles of CRA and FGSA: CRA supports stable risk recognition across sequential tasks, whereas FGSA translates risk awareness into selective control over federated model updates. Results under an alternative task sequence are provided in Appendix~\ref{app:results}.

\section{Conclusion}
\label{sec:conclusion}

In this paper, we propose Safe-FCL to enhance long-term safety in federated continual learning of LLMs. Safe-FCL integrates CRA, which preserves historical risk knowledge while adapting to emerging threats, and FGSA, which enables risk-aware aggregation through module-level safety control. Experiments demonstrate that Safe-FCL outperforms state-of-the-art methods, improving global-model Rule and MD-Judge scores by 0.41\% and 0.64\%, respectively, while increasing average risk recognition accuracy by 2.01\% and OOD accuracy by 4.93\%.

\section*{Limitations}

Although our empirical results demonstrate the effectiveness of Safe-FCL, several limitations remain. Our evaluation focuses on text-based federated continual learning under controlled malicious-client settings, while its effectiveness in multimodal and more complex adversarial scenarios, including adaptive attacks and coordinated malicious clients, remains unexplored. Moreover, as the number of clients and tasks grows, evaluation-branch optimization, detector adaptation, and historical feature retention may increase computational and storage overhead. Future work will investigate broader threat scenarios and more efficient trajectory evaluation and historical feature representations to improve the scalability and applicability of Safe-FCL.


\bibliography{custom,Safe_FCL_dataset_sources}

\input{Safe_FCL_Appendix_Compact}

\end{document}

%% file: Safe_FCL_Appendix_Compact.tex
\clearpage
\appendix

\setcounter{topnumber}{3}
\setcounter{totalnumber}{5}
\setcounter{dbltopnumber}{3}
\renewcommand{\floatpagefraction}{0.8}
\renewcommand{\dblfloatpagefraction}{0.8}
\makeatletter
\setlength{\@fptop}{0pt}
\setlength{\@dblfptop}{0pt}
\setlength{\@dblfpsep}{12pt}
\setlength{\@dblfpbot}{0pt plus 1fil}
\makeatother

\section{Dataset Sources}
\label{app:dataset-sources}

\subsection{Source Datasets}
Our data snapshot contains 49,765 question--answer (QA) pairs selected from 34 upstream repositories. Table~\ref{app:tab:sources} summarizes the data sources and their associated references, with related repositories grouped into 29 entries for presentation.

\input{Safe_FCL_Source_Table_34}

\subsection{Selection and Task Construction}

Our dataset contains 49,765 selected question--answer (QA) pairs across five safety domains and four partitions: 25,000 for LLM training, 3,570 for validation, 7,145 for testing, and 14,050 for auxiliary trajectory collection. These counts reflect the retained samples rather than the original sizes of the source datasets. 

The five safety domains are defined according to the primary targets and intended actions of requests, rather than directly adopting the taxonomies of source datasets. Related question families are grouped before domain assignment, data partitioning, and quota sampling. We categorize QA pairs into three types: benign requests with safe answers (A), harmful requests with safe answers (B), and harmful requests with harmful answers (C). Types A and B constitute the benign training pool, while type C forms the malicious pool.


\begin{table}[!htbp]
\centering
\footnotesize
\setlength{\tabcolsep}{4pt}
\renewcommand{\arraystretch}{1.15}
\begin{tabular}{@{}c p{0.64\columnwidth} c@{}}
\toprule
Task & Safety Domain & Role \\
\midrule
$\mathcal{T}_1$ &
Physical, public, and material safety & CL \\
$\mathcal{T}_2$ &
Privacy and information protection & CL \\
$\mathcal{T}_3$ &
Property, identity, and information deception & CL \\
$\mathcal{T}_4$ &
Digital security & CL \\
$\mathcal{T}_5$ &
Discrimination and interpersonal harm & OOD \\
\bottomrule
\end{tabular}
\caption{Safety task domains and evaluation protocols.
Each task contains 5,000 training, 714 validation,
and 1,429 test examples, with a benign-to-harmful
ratio of 7:3. CL denotes sequential continual
learning, and OOD denotes held-out evaluation.}
\label{tab:app_tasks}
\end{table}

\section{Experimental Protocol and Evaluation Details}
\label{app:experiments}

\subsection{Data Construction and Partitioning}
\label{app:data}

\noindent\textbf{Client Data and Threat Labels.}
Federated training data are partitioned among participating clients for each task. In our controlled threat setting, malicious-client identities are used to construct labeled training examples for the risk detector but are not provided to the server during federated aggregation. Thus, aggregation decisions rely on inferred risks rather than oracle knowledge of client identities.

\noindent\textbf{Risk-detector Examples.}
CRA uses update-space observations distinct from the examples used for LLM fine-tuning. These observations are constructed from benign and malicious local LoRA-update trajectories. Each observation consists of a feature vector extracted from cumulative parameter changes relative to a fixed task reference and a binary label indicating the update type. The labels are used for task-level detector adaptation, whereas subsequent aggregation relies only on predicted client risks.

\subsection{Evaluation Metrics}
\label{app:metrics}

\noindent\textbf{Continual Risk Recognition Metrics.}
Let $a_{k,j}$ denote the risk detector's classification accuracy (\%) on task $\mathcal{T}_j$ after adapting to task $\mathcal{T}_k$, where $j\leq k$. Average Accuracy is defined as:
\begin{equation}
\label{eq:app_aa}
\mathrm{AA}_k =
\frac{1}{k}\sum_{j=1}^{k}a_{k,j}.
\end{equation}

Average Forgetting (AF) measures changes in historical task accuracy relative to the reference accuracy $b_{k,j}$ specified by the evaluation protocol:
\begin{equation}
\label{eq:app_af}
\mathrm{AF}_k =
\frac{1}{k-1}
\sum_{j=1}^{k-1}(b_{k,j}-a_{k,j}),
\quad k\geq 2.
\end{equation}
AF retains signed values, with positive values indicating performance degradation and negative values indicating improvement relative to the reference.

New-task Accuracy (New ACC) measures the average accuracy immediately after learning each new task:
\begin{equation}
\label{eq:app_newacc}
\mathrm{New\ ACC} =
\frac{1}{K}\sum_{k=1}^{K}a_{k,k}.
\end{equation}

OOD Task ACC is evaluated on the held-out task using the detector frozen after the final continual learning stage.

\noindent\textbf{Global-model Safety Metrics.}
Let $G_k(q)$ denote the global model's response to query $q$ after learning task $\mathcal{T}_k$, and let $\mathcal{Q}_j$ denote the evaluation queries for task $\mathcal{T}_j$. For evaluator $m\in\{\mathrm{Rule},\mathrm{MDJ}\}$, the safety score is computed as:
\begin{equation}
\label{eq:app_safety_per_task}
S_{k,j}^{(m)} =
\frac{100}{|\mathcal{Q}_j|}
\sum_{q\in\mathcal{Q}_j}
\mathbf{1}\!\left[
\mathcal{J}_m(q,G_k(q))=\mathrm{safe}
\right].
\end{equation}

The stage-average safety score is calculated over all tasks learned so far:
\begin{equation}
\label{eq:app_safety_avg}
\overline{S}_{k}^{(m)} =
\frac{1}{k}\sum_{j=1}^{k}S_{k,j}^{(m)}.
\end{equation}

OOD Rule and MD-Judge scores are measured after federated fine-tuning on the held-out task, unlike OOD risk recognition accuracy, which is obtained without detector adaptation.

\section{Supplementary Results and Analysis}
\label{app:results}
This section presents supplementary experiments to evaluate Safe-FCL under different task sequences. We first report detailed per-task ablation results under the primary task sequence. We then consider an alternative task sequence, in which the four tasks are learned in the following order: digital security; privacy and information protection; property, identity, and information deception; and physical, public, and material safety. Discrimination and interpersonal harm remains the held-out OOD task. Under this alternative sequence, we evaluate global-model safety, component ablations, continual risk recognition, and OOD generalization.

\subsection{Per-Task Ablation Analysis of CRA and FGSA}
\label{app:ablation_primary}

\begin{table*}[!tp]
\centering

\caption{Per-task global-model safety ablation results (\%) under the primary task sequence. Bold and underlined stage averages indicate the best and second-best results per stage and metric, respectively (ties for best are bolded).}

\label{tab:safe_fcl_staircase_ablation}

\setlength{\tabcolsep}{5pt}
\renewcommand{\arraystretch}{1.09}

\resizebox{\textwidth}{!}{%
\begin{tabular}{@{}l c|ccccc|ccccc@{}}
\toprule
\multirow{2}{*}{Method} & \multirow{2}{*}{Task}
 & \multicolumn{5}{c|}{Rule}
 & \multicolumn{5}{c}{MD-Judge} \\
\cmidrule(lr){3-7}\cmidrule(l){8-12}

& & Task 1 & Task 2 & Task 3 & Task 4 & Avg.
  & Task 1 & Task 2 & Task 3 & Task 4 & Avg. \\
\midrule

\multirow{4}{*}{Safe-FCL w/o FGSA}
& T1 & 90.68 & -- & -- & -- & \textbf{90.68} & 92.54 & -- & -- & -- & \textbf{92.54} \\
& T2 & 93.47 & 92.54 & -- & -- & \underline{93.01} & 97.67 & 95.80 & -- & -- & \textbf{96.74} \\
& T3 & 84.62 & 91.38 & 77.39 & -- & 84.46 & 88.81 & 90.68 & 81.35 & -- & \underline{86.95} \\
& T4 & 93.71 & 91.38 & 92.31 & 93.71 & \textbf{92.78} & 95.10 & 95.57 & 92.77 & 99.07 & 95.63 \\
\midrule

\multirow{4}{*}{Safe-FCL w/o CRA}
& T1 & 89.74 & -- & -- & -- & 89.74 & 91.61 & -- & -- & -- & 91.61 \\
& T2 & 93.47 & 90.21 & -- & -- & 91.84 & 97.20 & 94.64 & -- & -- & 95.92 \\
& T3 & 85.78 & 92.31 & 75.76 & -- & \underline{84.62} & 89.74 & 91.61 & 81.35 & -- & \textbf{87.57} \\
& T4 & 94.41 & 93.01 & 89.51 & 90.44 & 91.84 & 95.34 & 95.80 & 93.24 & 98.37 & \underline{95.69} \\
\midrule

\rowcolor{oursblue}
& T1 & 90.44 & -- & -- & -- & \underline{90.44} & 92.31 & -- & -- & -- & \underline{92.31} \\

\rowcolor{oursblue}
& T2 & 94.41 & 92.07 & -- & -- & \textbf{93.24} & 97.67 & 95.57 & -- & -- & \underline{96.62} \\

\rowcolor{oursblue}
& T3 & 85.55 & 91.61 & 79.25 & -- & \textbf{85.47} & 88.81 & 91.84 & 82.05 & -- & \textbf{87.57} \\

\rowcolor{oursblue}
\multirow{-4}{*}{\textbf{Safe-FCL (Ours)}}
& T4 & 93.71 & 92.31 & 92.31 & 91.84 & \underline{92.54} & 95.10 & 95.10 & 94.64 & 99.07 & \textbf{95.98} \\

\bottomrule
\end{tabular}%
}
\end{table*}

Table~\ref{tab:safe_fcl_staircase_ablation} provides detailed per-task ablation results under the primary task order. After learning Task~4, Safe-FCL outperforms the variant without CRA on the previously learned Task~3 by 2.80\% in Rule and 1.40\% in MD-Judge, indicating the benefit of CRA for historical task safety. The effects of FGSA are more task-dependent. On the newly introduced Task~3, FGSA improves Rule and MD-Judge by 1.86\% and 0.70\%, respectively, whereas its Rule score on Task~4 is lower than that of the variant without FGSA. These results highlight the varying contributions of CRA and FGSA across individual tasks.

\subsection{Safety Performance under an Alternative Task Sequence}
\label{app:order1_safety}

\begin{table*}[!tp]
\centering

\caption{Global-model Rule and MD-Judge results (\%) under the alternative task sequence. Bold and underlined stage averages indicate the best and second-best results per stage and metric, respectively (ties for best are bolded).}

\label{tab:updated_rule_mdjudge_s1}

\setlength{\tabcolsep}{5pt}
\renewcommand{\arraystretch}{0.90}
\setlength{\aboverulesep}{1pt}
\setlength{\belowrulesep}{1pt}

\resizebox{\textwidth}{!}{%
\begin{tabular}{@{}l c|ccccc|ccccc@{}}
\toprule
\multirow{2}{*}{Method} & \multirow{2}{*}{Task}
& \multicolumn{5}{c|}{Rule}
& \multicolumn{5}{c}{MD-Judge} \\
\cmidrule(lr){3-7}\cmidrule(l){8-12}
& & Task 1 & Task 2 & Task 3 & Task 4 & Avg.
  & Task 1 & Task 2 & Task 3 & Task 4 & Avg. \\
\midrule

\multirow{4}{*}{FedAvg}
& T1 & 74.13 & -- & -- & -- & 74.13 & 80.19 & -- & -- & -- & 80.19 \\
& T2 & 64.57 & 60.37 & -- & -- & 62.47 & 66.43 & 64.57 & -- & -- & 65.50 \\
& T3 & 69.93 & 72.73 & 47.09 & -- & 63.25 & 69.46 & 72.03 & 42.19 & -- & 61.23 \\
& T4 & 72.73 & 80.65 & 43.36 & 34.03 & 57.69 & 73.19 & 80.19 & 38.69 & 32.17 & 56.06 \\
\midrule

\multirow{4}{*}{FLAME}
& T1 & 80.19 & -- & -- & -- & 80.19 & 82.98 & -- & -- & -- & 82.98 \\
& T2 & 73.43 & 72.73 & -- & -- & 73.08 & 74.59 & 76.22 & -- & -- & 75.41 \\
& T3 & 76.69 & 81.35 & 58.97 & -- & 72.34 & 77.62 & 82.75 & 54.08 & -- & 71.48 \\
& T4 & 73.43 & 82.28 & 47.55 & 35.43 & 59.67 & 73.19 & 83.22 & 42.89 & 33.33 & 58.16 \\
\midrule

\multirow{4}{*}{LASA}
& T1 & 81.12 & -- & -- & -- & 81.12 & 82.28 & -- & -- & -- & 82.28 \\
& T2 & 70.86 & 64.34 & -- & -- & 67.60 & 71.10 & 68.30 & -- & -- & 69.70 \\
& T3 & 74.83 & 75.52 & 48.25 & -- & 66.20 & 74.36 & 77.86 & 44.06 & -- & 65.43 \\
& T4 & 76.22 & 80.19 & 46.62 & 31.24 & 58.57 & 75.06 & 80.65 & 41.26 & 29.84 & 56.70 \\
\midrule

\multirow{4}{*}{Safe-FedLLM-Step}
& T1 & 85.08 & -- & -- & -- & 85.08 & 92.07 & -- & -- & -- & \underline{92.07} \\
& T2 & 87.88 & 87.18 & -- & -- & 87.53 & 92.54 & 94.64 & -- & -- & 93.59 \\
& T3 & 92.77 & 87.65 & 86.01 & -- & 88.81 & 93.47 & 92.54 & 89.74 & -- & 91.92 \\
& T4 & 88.11 & 93.71 & 79.25 & 69.00 & 82.52 & 88.81 & 94.17 & 81.35 & 74.59 & 84.73 \\
\midrule

\multirow{4}{*}{Safe-FedLLM-Client}
& T1 & 81.82 & -- & -- & -- & 81.82 & 86.71 & -- & -- & -- & 86.71 \\
& T2 & 84.38 & 80.65 & -- & -- & 82.52 & 87.65 & 90.68 & -- & -- & 89.17 \\
& T3 & 87.18 & 86.71 & 69.46 & -- & 81.12 & 87.88 & 89.28 & 68.76 & -- & 81.97 \\
& T4 & 83.92 & 92.54 & 55.71 & 45.45 & 69.41 & 83.92 & 91.84 & 51.98 & 45.22 & 68.24 \\
\midrule

\multirow{4}{*}{Safe-FedLLM-Shadow}
& T1 & 91.38 & -- & -- & -- & \underline{91.38} & 96.04 & -- & -- & -- & \textbf{96.04} \\
& T2 & 92.54 & 88.81 & -- & -- & \underline{90.68} & 95.34 & 98.60 & -- & -- & \underline{96.97} \\
& T3 & 92.07 & 93.01 & 91.38 & -- & \underline{92.15} & 93.01 & 96.04 & 95.10 & -- & \underline{94.72} \\
& T4 & 90.91 & 95.34 & 84.62 & 72.03 & \underline{85.73} & 92.31 & 96.74 & 89.98 & 82.52 & \underline{90.39} \\
\midrule

\rowcolor{oursblue}
& T1 & 92.31 & -- & -- & -- & \textbf{92.31} & 96.04 & -- & -- & -- & \textbf{96.04} \\

\rowcolor{oursblue}
& T2 & 94.64 & 89.98 & -- & -- & \textbf{92.31} & 95.80 & 98.83 & -- & -- & \textbf{97.32} \\

\rowcolor{oursblue}
& T3 & 92.54 & 94.41 & 93.47 & -- & \textbf{93.47} & 93.24 & 97.44 & 95.34 & -- & \textbf{95.34} \\

\rowcolor{oursblue}
\multirow{-4}{*}{\textbf{Safe-FCL (Ours)}}
& T4 & 91.61 & 95.57 & 86.71 & 78.55 & \textbf{88.11} & 93.71 & 97.20 & 91.14 & 83.68 & \textbf{91.43} \\

\bottomrule
\end{tabular}%
}
\end{table*}

Table~\ref{tab:updated_rule_mdjudge_s1} presents the global-model safety results under the alternative sequence. Safe-FCL achieves the highest average Rule scores across all four learning stages, while attaining the best or tied-best MD-Judge scores at each stage. After completing federated continual learning, Safe-FCL reaches average Rule and MD-Judge scores of 88.11\% and 91.43\%, outperforming current SOTA methods by 2.38\% and 1.04\%, respectively.

On the newly introduced Task~4, Safe-FCL achieves Rule and MD-Judge scores of 78.55\% and 83.68\%, exceeding current SOTA methods by 6.52\% and 1.16\%, respectively. These results further support the effectiveness of Safe-FCL in maintaining global-model safety across different task sequences.

\subsection{Component Ablation under the Alternative Task Sequence}
\label{app:order1_ablation}

\begin{table*}[!tp]
\centering

\caption{Global-model safety ablation results (\%) under the alternative task sequence, including OOD evaluation. Bold and underlined values indicate the best and second-best results per column, respectively (ties for best are bolded).}

\label{tab:safe_fcl_ablation_ood_s1}

\setlength{\tabcolsep}{6pt}
\renewcommand{\arraystretch}{1.15}

\resizebox{\textwidth}{!}{%
\begin{tabular}{@{}l*{5}{cc}@{}}
\toprule

\multirow{2}{*}{Method}
 & \multicolumn{2}{c}{Task1}
 & \multicolumn{2}{c}{Task2}
 & \multicolumn{2}{c}{Task3}
 & \multicolumn{2}{c}{Task4}
 & \multicolumn{2}{c}{OOD} \\

\cmidrule(lr){2-3}
\cmidrule(lr){4-5}
\cmidrule(lr){6-7}
\cmidrule(lr){8-9}
\cmidrule(l){10-11}

 & Rule & MD-Judge
 & Rule & MD-Judge
 & Rule & MD-Judge
 & Rule & MD-Judge
 & Rule & MD-Judge \\

\midrule

Safe-FCL w/o FGSA
 & \underline{92.07} & \underline{96.04}
 & \underline{91.96} & \textbf{97.32}
 & \underline{93.09} & \underline{94.64}
 & \underline{88.00} & \textbf{91.43}
 & \underline{88.34} & 92.54 \\

Safe-FCL w/o CRA
 & 91.14 & \textbf{96.74}
 & 90.09 & \underline{96.86}
 & 92.62 & \textbf{95.34}
 & 86.19 & \underline{90.27}
 & \textbf{93.94} & \underline{97.90} \\

\midrule

\rowcolor{oursblue}
\textbf{Safe-FCL (Ours)}
 & \textbf{92.31} & \underline{96.04}
 & \textbf{92.31} & \textbf{97.32}
 & \textbf{93.47} & \textbf{95.34}
 & \textbf{88.11} & \textbf{91.43}
 & \textbf{93.94} & \textbf{98.14} \\

\bottomrule
\end{tabular}%
}
\end{table*}

\begin{table*}[!tp]
\centering

\caption{Per-task global-model safety ablation results (\%) under the alternative task sequence. Bold and underlined stage averages indicate the best and second-best results per stage and metric, respectively (ties for best are bolded).}

\label{tab:safe_fcl_staircase1}

\setlength{\tabcolsep}{5pt}
\renewcommand{\arraystretch}{1.09}

\resizebox{\textwidth}{!}{%
\begin{tabular}{@{}l c|ccccc|ccccc@{}}
\toprule
\multirow{2}{*}{Method} & \multirow{2}{*}{Task}
& \multicolumn{5}{c|}{Rule}
& \multicolumn{5}{c}{MD-Judge} \\
\cmidrule(lr){3-7}\cmidrule(l){8-12}

& & Task 1 & Task 2 & Task 3 & Task 4 & Avg.
  & Task 1 & Task 2 & Task 3 & Task 4 & Avg. \\
\midrule

\multirow{4}{*}{Safe-FCL w/o FGSA}
& T1 & 92.07 & -- & -- & -- & \underline{92.07} & 96.04 & -- & -- & -- & \underline{96.04} \\
& T2 & 93.24 & 90.68 & -- & -- & \underline{91.96} & 95.80 & 98.83 & -- & -- & \textbf{97.32} \\
& T3 & 93.71 & 94.41 & 91.14 & -- & \underline{93.09} & 93.24 & 96.50 & 94.17 & -- & \underline{94.64} \\
& T4 & 91.14 & 96.04 & 87.41 & 77.39 & \underline{88.00} & 92.07 & 97.67 & 92.31 & 83.68 & \textbf{91.43} \\
\midrule

\multirow{4}{*}{Safe-FCL w/o CRA}
& T1 & 91.14 & -- & -- & -- & 91.14 & 96.74 & -- & -- & -- & \textbf{96.74} \\
& T2 & 92.77 & 87.41 & -- & -- & 90.09 & 95.34 & 98.37 & -- & -- & \underline{96.86} \\
& T3 & 93.71 & 91.84 & 92.31 & -- & 92.62 & 93.24 & 97.20 & 95.57 & -- & \textbf{95.34} \\
& T4 & 89.28 & 95.10 & 86.01 & 74.36 & 86.19 & 92.54 & 96.50 & 90.44 & 81.59 & \underline{90.27} \\
\midrule

\rowcolor{oursblue}
& T1 & 92.31 & -- & -- & -- & \textbf{92.31} & 96.04 & -- & -- & -- & \underline{96.04} \\

\rowcolor{oursblue}
& T2 & 94.64 & 89.98 & -- & -- & \textbf{92.31} & 95.80 & 98.83 & -- & -- & \textbf{97.32} \\

\rowcolor{oursblue}
& T3 & 92.54 & 94.41 & 93.47 & -- & \textbf{93.47} & 93.24 & 97.44 & 95.34 & -- & \textbf{95.34} \\

\rowcolor{oursblue}
\multirow{-4}{*}{\textbf{Safe-FCL (Ours)}}
& T4 & 91.61 & 95.57 & 86.71 & 78.55 & \textbf{88.11} & 93.71 & 97.20 & 91.14 & 83.68 & \textbf{91.43} \\

\bottomrule
\end{tabular}%
}
\end{table*}



Tables~\ref{tab:safe_fcl_ablation_ood_s1} and~\ref{tab:safe_fcl_staircase1} present the stage-average and per-task ablation results under the alternative sequence. After Task~4, Safe-FCL improves average Rule and MD-Judge scores by 1.92\% and 1.16\%, respectively, compared with the variant without CRA. On the previously learned Task~1, the corresponding improvements reach 2.33\% and 1.17\%, indicating the contribution of CRA to historical task safety. FGSA also provides improvements after Task~3, increasing average Rule and MD-Judge scores by 0.38\% and 0.70\%, respectively. On the newly introduced Task~3, these gains reach 2.33\% and 1.17\%.

The contribution of FGSA is particularly evident in OOD safety evaluation. Safe-FCL achieves Rule and MD-Judge scores of 93.94\% and 98.14\%, respectively, outperforming the variant without FGSA by 5.60\% on both metrics. These results suggest that combining continual risk awareness with module-level safety regulation helps maintain global-model safety across different task sequences, with notable benefits when adapting to an unseen domain.

\subsection{Continual Risk Recognition under an Alternative Task Sequence}
\label{app:order1_risk_recognition}

\begin{table*}[!tp]
\centering

\caption{Continual risk recognition performance (\%) under the alternative task sequence. Bold and underlined values indicate the best and second-best results per column, respectively (ties for best are bolded).}

\label{tab:safe_fcl_aa_af_ood}

\setlength{\tabcolsep}{6pt}
\renewcommand{\arraystretch}{1.15}

\resizebox{\textwidth}{!}{%
\begin{tabular}{@{}lccccccccc@{}}
\toprule

\multirow{2}{*}{Method}
 & \multicolumn{1}{c}{Task1}
 & \multicolumn{2}{c}{Task2}
 & \multicolumn{2}{c}{Task3}
 & \multicolumn{2}{c}{Task4}
 & \multirow{2}{*}{New ACC $\uparrow$}
 & \multirow{2}{*}{\shortstack{OOD Task\\ACC $\uparrow$}} \\

\cmidrule(lr){2-2}
\cmidrule(lr){3-4}
\cmidrule(lr){5-6}
\cmidrule(lr){7-8}

 & AA $\uparrow$
 & AA $\uparrow$ & AF $\downarrow$
 & AA $\uparrow$ & AF $\downarrow$
 & AA $\uparrow$ & AF $\downarrow$
 & & \\

\midrule

Safe-FedLLM-Step
 & 94.67
 & 85.00 & 20.67
 & 91.00 & 7.09
 & 79.62 & 17.00
 & 92.38 & 93.00 \\

Safe-FedLLM-Client
 & 98.33
 & 90.84 & 13.33
 & 91.67 & 10.00
 & 86.25 & 15.55
 & 97.08 & 88.33 \\

Safe-FedLLM-Shadow
 & \underline{99.63}
 & 85.38 & 28.80
 & 90.08 & 14.55
 & 91.59 & 10.55
 & 99.51 & \underline{94.40} \\

\midrule

Safe-FCL w/o CRA
 & \textbf{99.93}
 & \underline{98.10} & \underline{3.70}
 & \underline{99.66} & \underline{0.05}
 & \underline{99.36} & \underline{0.27}
 & \underline{99.56} & 89.93 \\

\rowcolor{oursblue}
\textbf{Safe-FCL (Ours)}
 & \textbf{99.93}
 & \textbf{99.99} & \textbf{-0.04}
 & \textbf{99.85} & \textbf{0.02}
 & \textbf{99.64} & \textbf{-0.01}
 & \textbf{99.62} & \textbf{95.87} \\

\bottomrule
\end{tabular}%
}
\end{table*}

Table~\ref{tab:safe_fcl_aa_af_ood} presents the continual risk recognition results under the alternative sequence. After Task~4, Safe-FCL achieves 99.64\% AA and $-0.01$\% AF, improving AA by 8.05\% and reducing AF by 10.56\% compared with current SOTA methods. Compared with the variant without CRA, Safe-FCL also improves final-stage AA by 0.28\% and reduces AF by 0.28\%, indicating better retention of historical risk recognition capabilities. Meanwhile, Safe-FCL achieves 99.62\% New ACC, exceeding current SOTA methods by 0.11\%, suggesting that historical risk knowledge retention does not compromise adaptation to emerging risks.

On the held-out OOD task, Safe-FCL achieves 95.87\% detection accuracy, outperforming current SOTA methods by 1.47\% and the variant without CRA by 5.94\%. Since the detector remains frozen during OOD evaluation, these results further support the ability of CRA to generalize learned risk recognition capabilities to unseen task domains without additional detector adaptation.

\subsection{OOD Global-Model Safety under an Alternative Task Sequence}
\label{app:order1_ood_safety}

\begin{figure}[!htbp]
  \centering
  \includegraphics[width=\columnwidth]{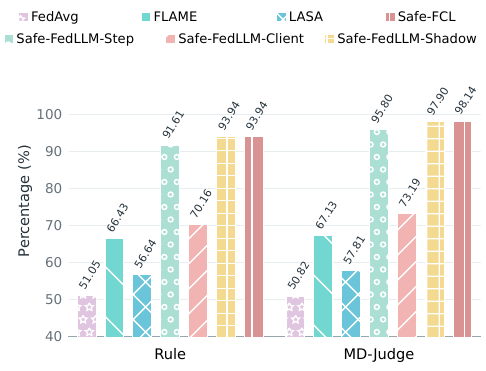}
  
\caption{OOD generalization of federated global models under an alternative task order. Rule and MD-Judge values are percentages.}

  \label{fig:ood-sequence1}
\end{figure}

Figure~\ref{fig:ood-sequence1} presents the OOD global-model safety results under the alternative task sequence. Safe-FCL achieves Rule and MD-Judge scores of 93.94\% and 98.14\%, respectively, matching the best Rule score among the compared baselines and improving MD-Judge by 0.24 percentage points over Safe-FedLLM-Shadow. Compared with Safe-FedLLM-Step, the corresponding improvements are 2.33 and 2.34 percentage points. Together with the primary-sequence results, these findings support the effectiveness of continual risk awareness and fine-grained safety aggregation in maintaining strong global-model safety under both evaluated task orders during adaptation to the held-out domain.

%% file: Safe_FCL_Source_Table_34.tex
\begin{table*}[!tp]
\centering
\begingroup
\color{black}
\hypersetup{citecolor=darkblue}
\footnotesize
\setlength{\tabcolsep}{4pt}
\renewcommand{\arraystretch}{1.10}
\begin{tabularx}{\textwidth}{@{}p{1.75cm}>{\raggedright\arraybackslash}X>{\raggedright\arraybackslash}p{3.5cm}r@{}}
\toprule
\textbf{Source IDs} & \textbf{Dataset or source group} & \textbf{Reference} & \textbf{Selected QA} \\
\midrule
S01 & Fenrir v2.1 (AlicanKiraz0) & \citep{sfcl_source01} & 108 \\
S02 & FalseReject & \citep{sfcl_falsereject} & 28 \\
S03 & HH-RLHF & \citep{sfcl_hhrlhf} & 471 \\
S04 & gpt-oss-20b red-teaming evals (ChestnutKurisu) & \citep{sfcl_source04} & 73 \\
S05 & Circuit Breakers & \citep{sfcl_circuit} & 456 \\
S06 & harmful-dataset (LLM-LAT) & \citep{sfcl_source06} & 2,323 \\
S07 & FinDER & \citep{sfcl_finder} & 600 \\
S08 & OpenEval (Open-Eval-Commons) & \citep{sfcl_source08} & 1,246 \\
S09, S11, S12 & BeaverTails and related releases & \citep{sfcl_beavertails} & 2,431 \\
S10, S13 & PKU-SafeRLHF and QA variant & \citep{sfcl_pkusafe} & 17,218 \\
S14 & CybersecurityQAA (Rowden) & \citep{sfcl_source14} & 382 \\
S15 & WildChat & \citep{sfcl_wildchat} & 356 \\
S16 & WildGuardMix & \citep{sfcl_wildguard} & 2,525 \\
S17 & WildJailbreak & \citep{sfcl_wildjailbreak} & 7,223 \\
S18 & Adversarial Arena & \citep{sfcl_arena} & 1,131 \\
S19 & FinQA & \citep{sfcl_finqa} & 48 \\
S20 & HarmfulQA & \citep{sfcl_harmfulqa} & 108 \\
S21 & StackExchange title/body QA (flax-sentence-embeddings) & \citep{sfcl_source21} & 953 \\
S22 & Gretel safety alignment en-v1 & \citep{sfcl_source22} & 996 \\
S23 & HarmAug & \citep{sfcl_harmaug} & 3,996 \\
S24 & LMSYS-Chat-1M & \citep{sfcl_lmsys} & 1,049 \\
S25 & cybersecurity\_qa (mariiazhiv) & \citep{sfcl_source25} & 17 \\
S26 & Privacy-Expert-Instructions (meeAtif) & \citep{sfcl_source26} & 245 \\
S27 & AEGIS2.0 & \citep{sfcl_aegis} & 165 \\
S28--S30 &
NVIDIA Nemotron safety datasets &
\citep{sfcl_source28,sfcl_source29,sfcl_source30} &
2,995 \\
S31 & cyberstrike-sft-120k (oyildirim) & \citep{sfcl_source31} & 4 \\
S32 & RAIL-HH-10K (responsible-ai-labs) & \citep{sfcl_source32} & 3 \\
S33 & toxic-dpo-v0.2 (unalignment) & \citep{sfcl_source33} & 226 \\
S34 & PolicyQA & \citep{sfcl_policyqa} & 2,389 \\
\midrule
\multicolumn{3}{l}{\textbf{Total: 34 recorded repositories, 29 displayed rows}} & \textbf{49,765} \\
\bottomrule
\end{tabularx}
\endgroup
\caption{
Complete coverage of 34 source repositories,
displayed in 29 rows, with 49,765 selected QA pairs
across all five domains and all four partitions.
The BeaverTails group includes BeaverTails,
PKU-SafeRLHF-10K, and PKU-SafeRLHF-30K
(S09, S11, S12).
The PKU-SafeRLHF group includes PKU-SafeRLHF
and PKU-SafeRLHF-QA (S10, S13).
The NVIDIA Nemotron group includes
Nemotron-3.5-Content-Safety-Dataset,
Nemotron-RL-Safety-v1, and Nemotron-SFT-Safety-v1
(S28--S30), contributing 130, 2,229, and 636
selected QA pairs, respectively.
All other repositories are listed individually.
References cite verified associated papers or
dataset repositories.
Grouping changes only the presentation,
not the recorded source identities.
}
\label{app:tab:sources}
\end{table*}